\documentclass[11pt, a4paper]{article}
\usepackage{graphicx}
\usepackage{caption}
\usepackage{subcaption}
\usepackage{booktabs}
\usepackage{array}
\usepackage{multirow}

\usepackage{url}
\usepackage[affil-it]{authblk} 
\usepackage{etoolbox}
\usepackage{lmodern}

\makeatletter
\patchcmd{\@maketitle}{\LARGE \@title}{\fontsize{18}{19.2}\selectfont\@title}{}{}
\makeatother

\AtBeginDocument{%
  \providecommand\BibTeX{{%
    \normalfont B\kern-0.5em{\scshape i\kern-0.25em b}\kern-0.8em\TeX}}}

\title{Measuring Data Collection Diligence for Community Healthcare}

\author{Ramesha Karunasena\textsuperscript{\rm 1}, 
Mohammad Sarparajul Ambiya\textsuperscript{\rm 2}, 
Arunesh Sinha\textsuperscript{\rm 1},
Ruchit Nagar\textsuperscript{\rm 2},
Saachi Dalal\textsuperscript{\rm 2},
Divy Thakkar\textsuperscript{\rm 3},
Dhyanesh Narayanan\textsuperscript{\rm 3},
Milind Tambe\textsuperscript{\rm 3}
\\
\textsuperscript{\rm 1}Singapore Management University, Singapore, \{rameshak,aruneshs\}@smu.edu.sg \\
\textsuperscript{\rm 2}Khushi Baby, India, \{sarfraz,ruchit,saachi\}@khushibaby.org \\
\textsuperscript{\rm 3}Google Research, India, \{dthakkar,dhyaneshn,milindtambe\}@google.com
}

\date{}
\begin{document}

\maketitle

\begin{abstract}
  Data analytics has tremendous potential to provide targeted benefit in low-resource communities, however the availability of high-quality public health data is a significant challenge in developing countries primarily due to non-diligent data collection by community health workers (CHWs). In this work, we define and test a data collection diligence score. This challenging unlabeled data problem is handled by building upon domain expert's guidance to design a useful data representation of the raw data, using which we design a simple and natural score. An important aspect of the score is relative scoring of the CHWs, which implicitly takes into account the context of the local area. The data is also clustered and interpreting these clusters provides a natural explanation of the past behavior of each data collector. We further predict the diligence score for future time steps. Our framework has been validated on the ground using observations by the field monitors of our partner NGO in India. Beyond the successful field test, our work is in the final stages of deployment in the state of Rajasthan, India.
\end{abstract}

\section{Introduction}
Community health workers (CHWs) play a vital role in health care systems, especially in the developing and under-developed parts of the world where CHWs provide last mile connectivity to reach rural communities. Among the many important contributions made by the these frontline CHWs, an important one is \emph{collection of population health data}. Such data is crucial for many tasks, such as health resource allocation, public policy decision making, health predictions, disease modelling, etc. However, even with extensive human infrastructure to enable data collection in \emph{developing countries}, some frontline community health workers (CHWs) are not able to deliver high quality data~\cite{columbiaglobal} due to various social and organization reasons. Data quality challenges have prevented the use of such data for downstream applications and when used, occasionally led to incorrect inferences and sub-otpimal outcomes. 
Poor data collection can be attributed \cite{pal2017changing,ismail2019empowerment} to various factors such as overburdened CHWs, lack of training on technology-mediated data collection tools, conflicting organizational incentives, low pay, lack of equipment, etc. We recognize the importance and challenges of CHWs~\cite{ismail2019empowerment,kumar2015projecting,yadav2017sangoshthi,world2013using} in providing access to medical care and health outreach to rural and marginalized communities. 
Field experts (including our NGO partners) struggle with fundamental design techniques of how to effectively measure the diligence of CHWs in terms of accurate data collection and reporting over time (measured in months), how to predict diligence for the next month, and how to design targeted interventions for improving data collection. 


We find that designing the data collection diligence score problem is a challenging problem. There are two main challenges: (1) absence of labeled data for indicators of poor and high diligence and (2) the diligence of a CHW is not expected to be of the same quality over time, which brings in a temporal aspect to the problem. While at first the problem may seem solvable by an unsupervised anomaly detection approach~\cite{eskin2000anomaly,chandola2009anomaly}, we found (and know from our NGO partners) that a large percentage of the reported data is suspected to be of poor quality, which rules out unsupervised anomaly detection techniques. Another approach, which is commonly used, is to utilize domain expert's knowledge of types of non-diligence and try to find such patterns in data. However, we found that such human designed non-diligence rules with fixed hard thresholds are too stringent as all CHWs violates some of these rules in every month and the fixed threshold fails to take the local context into account.

Our first technical contribution is to provide a framework for converting human designed rules of non-diligence detection into probabilities of non-diligence for each health worker. Typically, such rules measure 
some percentages in a time window, for example, one rule (BP rule) tracks the percentage of number of blood pressure readings in a health camp (over a month) that are all same, where 0\% is fully diligent and 100\% is fully non-diligent for this BP rule. We posit that a CHWs diligence must be measured \emph{relatively} with respect to the performance of other CHWs in the local area; such relative measurement implicitly accounts for the local context in which a health measurement task might be difficult due to factors such as lack of equipment. Further, we are interested in ranking the CHWs for the purpose of later intervention to improve diligence.
Thus, for each rule, we consider the relative position of a given percentage reading of a CHW in a month within the distribution of percentage readings of all CHWs in all months in the training data; this relative position is converted to a probability of non-diligence for the CHW for that rule in a time window. Recognizing that such rules may change (added or deleted too) over time our framework is capable of modification of rules via a configuration file.
Then, the multiple rules provide a multi-dimensional vector of non-diligence probabilities for every month.
We call these health-worker specific probability vectors as their \emph{behavior}. However, a multi-dimensional vector does not allow ranking CHWs according to non-diligence.

Our second technical contribution includes a definition of a scalar non-diligence score for each CHW for each month as the Euclidean norm of the multi-dimensional behavior vector described above.
Then, we predict this scalar score using the past six months of a CHW's scalar scores as features in a linear regression predictor. We observed that, given limited data size, linear regression outperforms other complex models such as neural networks. The ranking enabled by scalar scores allows for intervention using limited resources in future.

Our final technical contribution is to provide fine grained explanation of the past behavior of each health worker by clustering the behavior vectors into multiple clusters and interpreting each cluster using the cluster center. We find that k-means with three clusters works best in our specific application. 
We also perform this interpretation of the clusters in three varying information levels, where a high level abstract explanation is as simple as saying diligent or non-diligent behavior and low level explanations provide details of which rule the CHW is not doing good at. The fine-grained explanation provides insights about the broad types of non-diligence exhibited in past by a CHW for targeted intervention for improving diligence in future.

Our \emph{main highlight} of this work is a field test, where our diligence score and behavior explanation of each CHW was validated against on-field observations made by field monitors (from our partner NGO) who observed the CHW on site in the state of Rajastan, India. Overall we observe that 67\% of our predictions match with the observations of the field monitors; moreover, in additional 17\% cases our tool correctly identifies non-diligence that the field monitors miss out. Thus, we achieve a 84\% correct non-diligence detection in the field. In addition, we are also able to precisely explain the reasons for not matching in the remaining 16\% cases. 
Based on this field test, we have started the deployment process of this software in the state government controlled server for use by our NGO partner. We firmly believe that our data collection diligence score, prediction of the same and the behavior explanations coupled with well-designed apriori and post-hoc intervention, such as more informed training or nudges and auditing, will over time result in improved data quality in the setting we are concerned with. Overall, our system can potentially lead to positive health outcome for millions of people throughout the developing world.

\section{Related Work}
\label{sec:related}
There is a long history of dealing with data quality issues in machine learning. Many of these approaches focus on scenarios where the data is noisy or adversarial corrupted 
but the amount of such noise or corruption is small. This low perturbation assumption naturally leads to unsupervised anomaly detection based techniques~\cite{eskin2000anomaly,chandola2009anomaly,xiong2011hierarchical,mccarthy2013towards,xiong2011group,yu2015glad,naidoo2020unsupervised}. However, as stated earlier, poor quality data is not an anomaly in our problem; in fact, majority of the data is poor quality due to non-diligence in collecting data. Also, with no label, supervised anomaly detection~\cite{gornitz2013toward} cannot be used. Further, distinct from these prior work, our goal is not to measure the data quality in aggregate but to measure the diligence of the data collector, improving which would naturally lead to better quality data. 

More closely related to our work is a work that directly focuses on measuring anomalous data collection by health workers~\cite{mccarthy2013towards}. However, this work also treats poor data collected by a health worker as an outlier, which is not true in our problem. In experiments, we perform a baseline comparison with anomaly detection for sake of scientific completeness. In not so closely related work, detecting human behavior in context of fabrication~\cite{birnbaum2013using} has been studied using behavior data from a controlled user study; we do not possess behavior data of health workers and it is quite infeasible to collect such data.



Clustering has been used in past work to detect fraud in social networks~\cite{cao2014uncovering}, in internet advertising~\cite{tian2015crowd}, in healthcare billing~\cite{massi2020data,liu2013healthcare}. However, as far as we know, neither clustering nor cluster explanations have been used in a problem context as ours. Moreover, our problem is not one of fraud, which is easier to define, but, one where we rely on human input to define non-diligence. Further, we explain the clustering result to yield better outcomes overall.

There are other kinds of data quality issues considered in machine learning. These include data missing over a large part of the feature space~\cite{lakkaraju2017identifying}, adversarially corrupted data~\cite{papernot2016towards,biggio2018wild}, and imputation for missing data in healthcare~\cite{hu2017strategies}. Our concern in this work is orthogonal to all these other issues. Also, there is work on incentivizing workers in crowd-sourced data labelling to obtain correct labels~\cite{singla2013truthful,mao2013volunteering,radanovic2016incentives,shah2015approval}; our context is very different: the data collection task is extremely arduous and the reasons of erroneous data collection are very varied and nuanced (as mentioned in introduction).

\section{Problem Description}
\label{sec:problemdesc}

\begin{figure*}[t]
\begin{subfigure}[b]{0.52\textwidth}
\includegraphics[width=\textwidth, height=5.5cm]{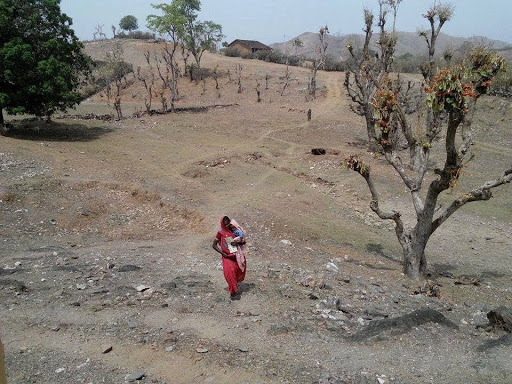}
\caption{A mother and an infant, walking several kilometers on foot to visit the only health camp in the month}
\end{subfigure}~~
\begin{subfigure}[b]{0.44\textwidth}
\includegraphics[width=\textwidth, height=5.6cm]{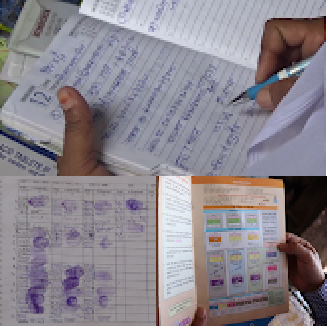}
\caption{Paper records filled by health workers at camps, which are later entered into a web portal manually} 
\end{subfigure}
\caption{Challenging circumstances for data collection in the rural areas of Rajasthan, India}
\label{fig1}
\end{figure*}

In this work we collaborate with a NGO, whose main mission is to motivate and monitor the health care of mothers and children to \emph{the last mile}. Using human-centered-design, they have developed a technology platform for longitudinal tracking of maternal and child health using a combination of a mobile application for healthworkers, a Near Field Communication enabled smart health card and automated local-dialect voice reminders for beneficiaries, and automated high-risk and dropout reports over WhatsApp for health officials and teams. They have deployed their platform to track the health of over 40,000 mothers and infants at the last mile in Rajasthan in India in partnership with the district government. 

The health workers, called ANMs (Auxiliary Nurse Midwives), are employed by the state government and are responsible for screening pregnant women and infants at these village-based maternal and child health camps. Each ANM provides antenatal care checkups and infant checkups (including immunizations) on four to five fixed camp sessions, each in a separate village, typically on each Thursday of the month. On average, an ANM caters to approximately 200 beneficiaries across these villages. An ANM also holds other responsibilities related to other health vertical programs which she balances with provision of reproductive and child health services. The data collection task is difficult and was paper based (see Fig.~\ref{fig1}). The NGO has enabled a streamlined digital data collection service for ANMs to collect data from patients in health camps, reducing some of the burden of the ANMs.
The NGO has made available datasets that correspond to health camps that have been conducted in the past by ANMs. 

The NGO's data science team also provided a set of 11 heuristic rules and associated thresholds, which is what they were using to assess the quality of data collected---in particular, whether some health records are indicative of lack of diligence on the part of the ANM when screening a patient. Examples of non-diligence include: filling in test results in the absence of conducting the test, rounding results when more precise data is available, failing to record data fields that are non-mandatory, and recording test results for certain patients but reporting that the same test equipment is not available for other patients in the same camp. Analysis of the state's database (ANMs send data to state database either via the NGO's system or via paper records) and observations from the field have shown major gaps in data fidelity. Less than 50 percent of expected maternal and child deaths are reported in the state's database.
Frequencies of normal urine studies are reported as high, when in many cases, due to lack of a private bathroom at the health camp, women are not asked to produce a urine sample for the test. Frequencies of blood pressure values in the state database show evidence of rounding, with low reporting of hypertension, which is expected at a certain baseline frequency during pregnancy. Even the number of actual blood pressure examinations done has been called into question~\cite{columbiaglobal}.

\textbf{Dataset description:} The data from the NGO is for 74 ANMs. Every ANM conducts health camps at particular locations at regular intervals in her (ANMs are females) jurisdiction. Pregnant women are advised to get at least four check-ups in the course of the pregnancy and they visit the nearest camp, when due for a check-up, for the same. The data recorded is quite extensive ranging from basic health parameters such as blood pressure and sugar levels to pregnancy-specific parameters such as fundal height and baby position. All data about the beneficiaries who visit the health camp was de-identified and anonymized. Some health test happen at every camp and others (like HIV status) are conducted once in the pregnancy at a primary health center. Other population wide statistics, such as new-born child mortality rate, are also recorded. The data provided was initially from 01 January 2020 to 31 January 2021. Maternal and child health camps were suspended in April 2020 due to the COVID-19 outbreak. A data drop from February 2021 to March 2021 was used for field test.

\textbf{Problem statement:} The long term goal of the NGO is to use the data collected for AI based support for better health care. However, the NGO knows from experience as well as confirmation from other NGOs that the data collected by ANMs is often not accurate. Thus, there is a need to identify non-diligent ANMs. Concretely, this requires solving three problems: (1) how to quantitatively measure non-diligence per ANM (2) how to explain the behavior of ANM over time and (3) how to predict non-diligence per ANM in order to enable apriori intervention.

\textbf{Issues with data:} In our working with the data-set we iteratively discovered a number of issues with the data. As we describe later, these issues are critical as our first iteration of the solution was misled by these data issues. The first main issue is that some ANM's data had few patients per health camp; however, it was not known if these ANMs really had few patients or were they not entering data into the digital system. During the course of the project, it was found out (by comparing with the state database) that these ANMs were not entering a large chunk of their data into the digital system and as a consequence we decided to filter out such ANMs. This led to filtering out of eight ANMs. 

The NGO also maintained a ranking of the ANM based on input of the field team of the NGO. While the ranking stays the same over time, our initial attempt was to treat this ranking as labels of diligence. However, over the course of the project we found two facts: (1) the rankings often contradict with what the data says (in terms of rules), for example, highly ranked ANMs were found to be entering all blood pressure values as same and (2) the ranking correlated with the number of the patients of the ANMs, probably reflecting a subjective view of which ANM is diligent. Based on these findings and in discussion with the field team of the NGO, we mutually decided to not use this ranking.

\section{Methodology}
\label{others}
\begin{figure*}
    \centering
    \includegraphics[width=\textwidth]{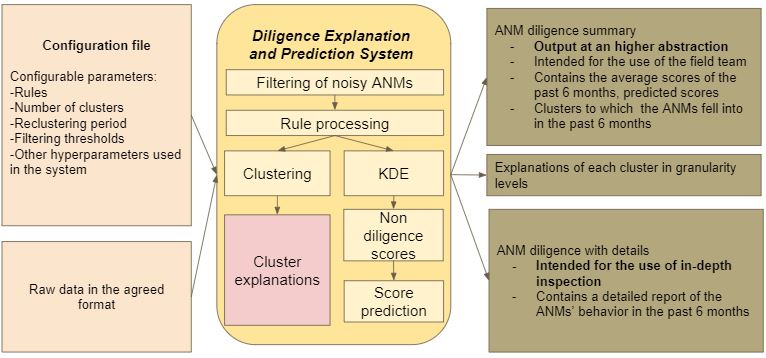}
    \caption{System diagram}
    \label{fig:systemdiag}
\end{figure*}

We describe our approach in three distinct parts: first is a definition and multi-dimensional score of non-diligence per ANM, second is a scalar non-diligence score for each ANM and prediction of the same, and finally generation of behavior explanations. A summary view of the system is shown in Fig.~\ref{fig:systemdiag}. 
Our code for the network is available publicly at \cite{gitsource}.

\subsection{Defining Non-diligence}
\label{sec:definingND}
Our first main idea is to obtain a probability of non-diligence per ANM corresponding to each of the domain-expert rules provided by the NGO using the data generated by each ANM. There are 11 rules as of now (March 2021) and additional rules can be added by defining them in the configuration file of the system. These 11 rules are of two types (1) rules that track known non-diligent phenomenon (5 in number) and (2) rules that track contradiction in data records (6 in number). All rules specify a percentage of something and we know which extreme (0\% or 100\%) or a defined range corresponds to diligence. For reasons of required confidentiality of these rules, we do not list all the actual rules used in the work. As a running example we will use two rules throughout this paper:

\textbf{Running examples}: \emph{Known non-diligence rule}: percentage of blood pressure readings in a month that are 120/80 or 110/70. We know that higher percentage corresponds to non-diligence. We call this the BP rule.
\emph{Contradiction rule}: When an ANM has entered ``No equipment available'' for few patients yet records fetal heart rate readings for other patients, it is a contradiction in fetal heart rate reading. We get the percentage of ``No equipment available'' records out of the total number of records in a month. 0\% means no contradiction. Thus,  lower percentage means more diligent and higher percentages are non-diligent readings.

\begin{figure}
    \centering
    \includegraphics{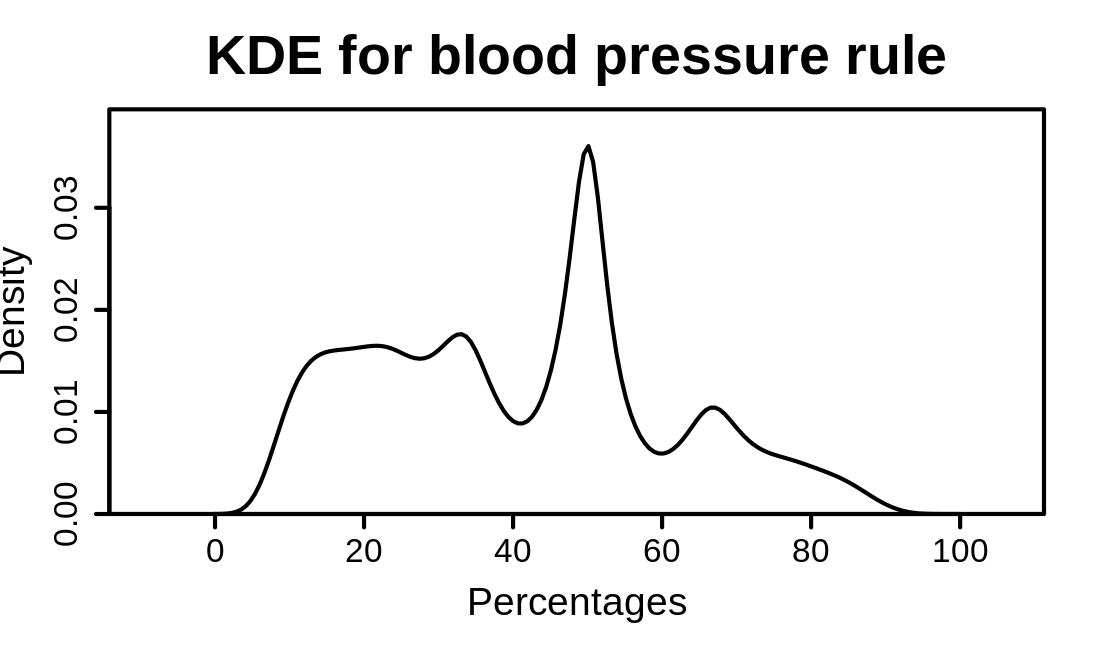}
    \caption{Kernel Density Estimation plot for the blood pressure rule after removing extremes}
    \label{fig:kde}
\end{figure}

\textbf{Handling Rules:}
We describe our approach here using the BP rule stated in the running example; rest of the rules are similar or flipped where 0\% is diligent. 
For each month $t \in \{1,\ldots, T\}$ in the training data, 
we compute the percentages as stated in the BP rule for each ANM $i \in \{1, \ldots, n\}$. This gives us the list of percentages for the BP rules as $\{P_{i,t}\}_{i\in \{1, \ldots, n\},t \in \{1,\ldots, T\}}$. Then, we filter out the percentages that are exactly 0\% or 100\% from this set $\{P_{i,t}\}$ - as these extremes are for sure diligent or not. We treat the remaining percentages in $\{P_{i,t}\}$ as samples from a probability distribution and plot a probability density distribution using Kernel Density Estimation (KDE). See Figure~\ref{fig:kde} for an example KDE plot for the BP rule. 

Then, given the percentage for the BP rule, say $P_{i,t}$, in a month $t$ for an ANM $i$, the probability of non-diligence $p$ is the probability mass between $(0,P_{i,t})$. Clearly as $P_{i,t}$ increases, the probability of non-diligence is increasing and is exactly 1 when percentage is 100. 
For other rules, where non-diligence with probability 1 is at the 0\% extreme,  the probability of non-diligence $p$ is the probability mass between $(P_{i,t},100)$. 

It is worth noting that our measure of probability of non-diligence for a rule can be viewed as a measure of the current month's performance of an ANM relative to all the past performances of all the ANMs. This relative scoring is important as certain measurement tasks are inherently difficult due to local factors such as lack of equipment, drop out of patients, etc. Thus, an absolute threshold is meaningless without taking into account the difficulty of the health data collection task. Our relative scoring is analogous to relative grading in university courses where the relativeness naturally takes into account the hardness of the underlying test.

\textit{Time window choice of one month:} The size of the time window is an important consideration. The ANM behavior varies a lot from one health camp to another, which is why we do not choose a time window as one health camp (one week). We found that a time window of one month (four to five health camps) exhibits more stable behavior. Applied over all the rules this yields 11 non-diligence probability values, one for each rule, for each time window of one month. 



\textbf{Filtering of noisy ANMs}:
As stated earlier, we ignore ANMs with less data. Keeping the data of these ANMs can mislead the estimation of the distribution and hinder the measurement of diligence. 
We filter out the ANMs according to two thresholds (both can be configured using the configuration file): 
(1) ANMs who recorded patients below a given threshold in each month and
(2) ANMs who recorded a total number of patients below a given threshold in a year. 
Out of 74 ANMs in the training data set, 66 ANMs were selected after filtering out the noisy ANMs.

\textbf{Vector valued score of ANM behavior}: The above approach provides 11 non-diligence probability values (vector of probabilities $\vec{b}$) to measure each ANM's behavior at any given point in time by using past one month of data. 
Next in Section~\ref{sec:scores}, we describe our approach to obtain a single scalar valued non-diligence score for each ANM, which is required for the purpose of ranking the ANMs.



\subsection{Non-diligence Score and Prediction}
\label{sec:scores}
The ideal behavior vector for an ANM is $\vec{0}$, meaning zero probability of non-diligence for every rule. Thus, given a vector $\vec{b}$, a natural score  is its distance from $\vec{0}$, that is, $||\vec{b}||_2$. This score is interpretable and by definition explicitly gives equal importance to all rules. We define this as the \emph{data collection diligence score}
We aim to predict this diligence score of an ANM in the dataset. 
For prediction we use a simple linear regressor, since our data is limited. We use the diligence scores of the past six months for each ANM as features to predict the  diligence score for the next month.

\subsection{Clustering and Generating Interpretations}
\label{sec:clust_scores}
The past scores by themselves provide only a view of the relative diligence of an ANM. Further, these score do not broadly identify patterns of non-diligence. In this section, we use clustering of raw percentage for every rule to obtain broad types of non-diligence and also interpret the clusters these to generate description of the ANM's past behavior.

After obtaining the percentage vectors (11 dimensional corresponding to each rule) for every month and every ANM in the training data, we cluster these vectors into three clusters using $k$-means. We set the number of clusters as three, since a clear separation 
(guided by the elbow method~\cite{5961514})
was observed when $k$ was chosen as three. However, $k$ is configurable in the system using the configuration file in anticipation of changed patterns in future. Observe that we use raw percentages when clustering and not the non-diligence probabilities. This is because the non-diligence probabilities are relative measures of performance and we wish our clusters to capture absolute ANM behavior in order to provide the correct targeted messages in the intervention stage (see also the description of least important rule later in this sub-section).

Once the clusters are determined using the current data, the clusters are fixed for n number of months, where n is configurable in the system. That is, no future data is used to update the clusters for n months and reclustering is done only after n months. This is to ensure that we measure the future behavior vectors within a static frame.  

\textbf{Generating behavior interpretation}: 
We generate interpretations for each cluster and call this as the \emph{behavior interpretation}.
Note that we do not have labels for ANM's behavior, but the nature of our constructed features (percentages for each rule) readily allows us to interpret the meaning of the clusters. Denote the 11 dimensional cluster center of cluster $k$ as $\langle P^1_k, \ldots, P^{11}_k\rangle$. We convert the cluster centers (which is in term of percentages) to non diligence probabilities $\langle p^1_k, \ldots, p^{11}_k\rangle$ using the KDE approach we described in Section~\ref{sec:definingND}. We call the non diligence probabilities of the cluster centers as \emph{cluster center diligence vector}


We use the cluster center diligence vector to generate interpretations at different granularity levels. At lower levels, we provide information at a higher abstraction. For example, at level 0, we output whether the cluster is generally good or bad. More precisely, we compare the average ($\sum_{r=1}^{11} p^r_k / 11$) non-diligence probabilities indicated by the cluster center; based on whether this average for a cluster is clearly is greater than that of the another cluster, we tag the clusters as non-diligent (bad) and diligent (good). 

For a more fine-grained interpretation of clusters, we partition the rules into more discerning and less discerning. We do so by computing the standard deviation of the percentages $\{P_{i,t}\}$ of all ANMs and time windows for each of the 11 rules. We create three partitions of the rules based on these 11 standard deviation values and call these the most (highest stddev), less, and least (lowest stddev) important rules; the thresholds for standard deviation are chosen by tuning. The least important rules are those where (mostly) percentages for all months and all ANMs are very close. That is, either all ANMs perform diligently or all ANMs perform non-diligently in these least important rules; the cause of this might be due a very difficult or very easy underlying health data collection task. This also reveals why using non-diligence probability is not correct for clustering as the probability score will assign relative scores from 0 to 1 to the ANMs even with small difference in the percentage measured for these rules. While this is fine for relative scoring, this is not fine for understanding behavior patterns, for example, ANMs with very similar performance on a rule might receive very different messages of how to improve if just the scores are used for this rules. 
The most important rules shows higher variance of behavior and are considered important as intervention for these rules inform us that since it is possible for some ANMs to do good with respect to these rules, hence other ANMs can be encouraged to do better.

At level 1, for each cluster we output diligent or not in \emph{each of the most important rule}. We do so by saying that if the cluster $k$ center non diligence probability for a rule $r$ is more than the average ($\sum_{k=1}^{3} p^r_k / 3$) cluster center non diligence probability then this cluster $k$ is interpreted as capturing non-diligent behavior for rule $r$. At level 2, we provide the same type of explanation for each of the less important rules. We do not provide interpretation for the least important rules, since there is no significant difference in the performance of the ANMs for these rules among the three clusters. 

Given the cluster to which an ANM belongs in a given month, the field team can identify the diligence of the ANM for every rules using our behavior descriptions at different granularity levels. This also aids in designing future interventions accordingly. The history of clusters of an ANM also provides an idea about her behavior trend over time.

\section{Results}
\label{sec:results}

We present our experimental results in this section. The data-set was already described earlier in Section~\ref{sec:problemdesc}. An additional detail is that, the health camps were disrupted due to the COVID caused lockdown in India and only very few ANMs conducted camps. Thus, data in April 2020 were dropped, since they might not be reliable. 

We process the data as we described earlier. We split the data set we have into training and test data, where the split was according to time due to the sequential (time series) nature of data. Our training data was from January 2020 to November 2020 and test data from December 2020 to January 2021. An on-field observation exercise was initiated by field monitors of the NGO in February-March 2021, which is used to validate our results.

For KDE, we use the kde1d library in R (invoked from python) with the an adaptive Gaussian kernel bandwidth~\cite{sheather1991reliable}.
All our code is written in python and R and ran on a cloud infrastructure. We start by presenting results from a baseline method.

\subsection{Baseline Methods Results}

We tried two baselines. The first is a simple heuristic baseline that uses the 11 rules provided by the NGO with fixed thresholds percentage. For example, for the BP rule a threshold of $70\%$ of all blood pressure readings in a health camp being exactly same was used to tag ANMs as diligent or not. Note that these rules do not provide a score, so are not very desirable to start with. Moreover, these rules performed very poorly---the outcome we obtained was that either (1) no ANM was tagged as non-diligent when the rules were used in an AND manner and (2) all ANMs were tagged as non-diligent when the rules were used in an OR manner.

The other baseline that we try is anomaly detection. We used a popular variational auto-encoder based anomaly detector~\cite{An2015VariationalAB}. We pre-processed our training data over time similarly as for clustering by using one month time window and processing all rules as percentages. We did not convert the raw percentage to probabilities. We trained the anomaly detector on the training data and then tested how the non-diligence score differs for the set of ANMs tagged non-diligent vs the set of ANMs tagged diligent by the anomaly detector in the test data. We use the norm score to compare these two sets as the norm score is interpretable. The results for the two months in test data is shown in Table~\ref{tab:vae}. ANMs tagged as non diligent should have a high non diligence score than ANMs who are not tagged in each month, which is not the case here.
As stated earlier, this is likely because non-diligence is not an outlier and unsupervised anomaly detection is based on the outlier assumption.
\begin{table}[t]
\centering
\caption{Mean values of non-diligence scores of the ANMs tagged non-diligent and non-tagged by the anomaly detector}
\label{tab:vae}
\begin{tabular}{ p{4cm}  p{2.2cm}  p{2.2cm}  }
\toprule
\bfseries  & \bfseries December 2020 & \bfseries January 2021\\
\midrule
Tagged ANMs & 1.8499 & 1.8391 \\
Non-tagged ANMs & 1.8537 & 1.7898 \\
\bottomrule
\end{tabular}
\end{table}

\subsection{Clustering Results}

As described earlier our clustering done on the raw percentages provided three clusters, for which we present the three cluster centers (in percentages) in Table~\ref{clusters}. The rules have not been individually named in order to preserve secrecy. Rules 1-5 are known non-diligence rules and 6-11 are contradiction rules. We obtain the diligence probabilities of the cluster centers and at level 0 of interpretation  the cluster with higher average probability of non-diligence is the non-diligent behavior cluster. Generally, cluster 0 is good in all rules (good cluster); cluster 1 is bad in most known non-diligent rules (including BP rule, which is the 1st rule mentioned in the Table~\ref{clusters}) and good in contradiction rules; and cluster 2 is bad in contradiction rules, but good in some known non-diligent rules like BP rule. Thus, at level 0 interpretation both cluster 1 and cluster 2 are non-diligent clusters. More levels of cluster interpretations are generated as explained in Section~\ref{sec:clust_scores}. In particular, four rules were found to be least important: rules 2,6,7,8, for which no explanations are generated. Rule 3 and 4 barely crossed the threshold for being least important and on closer inspection we do find ANMs with good behavior for these rules in cluster 0.

\begin{table*}
\centering
\small
\caption{Cluster centers (Rules are anonymized and numbered 1 to 11)}
\label{clusters}
\begin{tabular}{l  p{0.5cm}  p{0.5cm}  p{0.5cm}  p{0.5cm}  p{0.5cm}  p{0.5cm}  p{0.5cm}  p{0.5cm}  p{0.5cm}  p{0.5cm}  p{0.5cm} }
\toprule
\bfseries Cluster & 1 & 2 & 3 & 4 & 5 & 6 & 7 & 8 & 9 & 10 & 11 \\
\midrule
\bfseries Cluster 0 & 10.50 & 0.94 & 97.43 & 86.04 & 
12.67 & 0.15 & 0.72 & 0.29 & 1.46 & 7.33 & 2.31 \\
\bfseries Cluster 1 & 70.36 & 0.12 & 99.90 & 
88.54 & 3.75 & 2.20 & 2.02 & 1.56 & 8.09 & 4.40 & 4.90 \\
\bfseries Cluster 2 & 11.27 & 1.02 & 99.19 & 89.97 & 
10.44 & 8.11 & 2.24 & 7.20 & 59.97 & 20.04 & 27.44 \\
\bottomrule
\end{tabular}
\end{table*}

\subsection{Prediction Results}
\begin{figure}[t]
\centering
\includegraphics[width=0.45\textwidth]{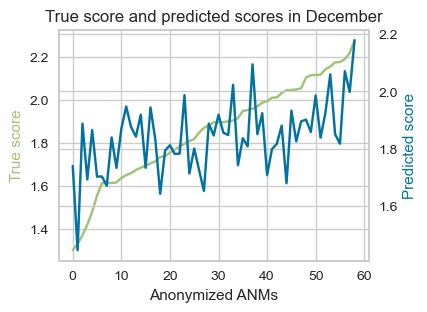}
\includegraphics[width=0.45\textwidth]{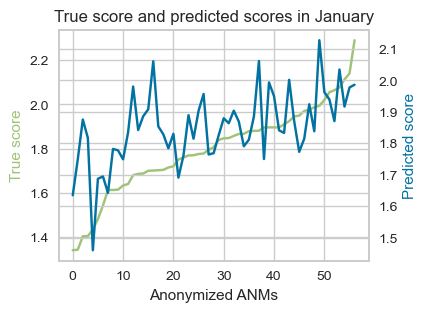}
\caption{True scores and predicted scores per ANM distribution in December 2020 and January 2021 (x-axis are anonymized ANM ids)}
\label{fig:scores_dist}
\end{figure}

We possess only small amount of data spanning about one year of and 66 valid ANMs. Our initial attempts was to use shallow neural networks. Although neural network had reasonable accuracy, MSE and R2 scores, they were not in agreement with the field monitors. Here, we have used a simple linear regressor and the results are presented in Table~\ref{predictor} which produces quantitatively similar results as neural networks but agrees more with filed monitors. The distribution of the predicted scores and true non-diligence scores of the test data is shown in Figure~\ref{fig:scores_dist} and correlation (Pearson) of the prediction and true score is on average 0.54 (correlations can be between [-1,1] with 0 denoting no correlation).

We do not observe very high R2 scores or perfect correlation, but the scores when paired with cluster explanations provide a good measure of diligence, which is validated in the field (explained in the subsequent section). 

\begin{table}[t]
\centering
\caption{MSE and R2 score of the predictor for test data}
\label{predictor}
\begin{tabular}{ p{1.5cm}  l  l  }
\toprule
\bfseries Metric & \bfseries December 2020 & \bfseries January 2021\\
\midrule
MSE & 0.0401 &  0.0296 \\
R2 & 0.2575 & 0.2624 \\
\bottomrule
\end{tabular}
\end{table}


\subsection{Field Test Results}
We have validated our results using the observations made in a field monitoring exercise carried out in February-March 2021. The field monitor were sent out to randomly chosen ANMs for one week each for a total of 37 observations. Each chosen ANMs was observed for one week and the whole exercise spanned four weeks. The field monitors are employed by the NGO and are known to the ANMs as the monitor help the ANMs with the digital equipment as and when required. The field monitoring task is itself very arduous as it involves travelling to many remote parts. The field monitors were provided a list of questions relating to the ANM performance (a summary of the questionnaire is in Appendix~\ref{questions}) and they recorded responses to these questions based on their observation. The field monitors' observations may also not be perfect due to various human factors involved such as mistakes, observation only for limited times, diligence of field monitors, etc. In our evaluation of the monitors recording, we discard monitor's observations for one ANM due to contradictory answers for the observation questions for this ANM. Our NGO partner's infer (manually) whether an ANM is diligent in a week based on the overall evaluation of the field monitor, which is what we compare with the output of our AI system.

For this field test, we need to put the AI system's scores into discrete buckets of diligent and non-diligent. We treat top 30\% ANMs (top 20) as diligent, bottom 55\% ANMs (bottom 36) as non-diligent and the middle 15\% (10 ANMs) as ANMs whose performance is determined by past trend in score and past clusters. Among the middle 10 ANMs, we mark those as diligent who have a non-increasing trend in score and belong to good clusters in past months. We deliberately use our clusters for putting medium score ANMs into buckets in order to extract the maximal information from the AI system as possible. Moreover, in the planned intervention in future we plan to be conservative and commend the top 10 ANMs and encourage the bottom 20 ANMs to perform better. The clusters will help in targeted encouragement about which aspects to improve on.

Our NGO partners find that our AI system's results are quite useful. Results are provided in Table~\ref{observations} for the 36 observations that we considered finally. We see a 67\% agreement rate with the monitors' observations. Moreover, for 17\% cases when AI predicts non-diligent, we can analyze the actual performance after-the-fact with the data collected by the ANM during the observation period. We find that the AI was correct in all these cases as there is evidence of non-diligence in the data collected during the observation period (see Appendix~\ref{field} for these details). Thus, AI's success rate is about 84\%. 

Further, for other cases in which AI output non-diligent, we find that some of these cases of non-diligence could not have been detected from given data. For example, ANM filling up all data for urine test from past experience without actually conducting any test as well as discovery of the fact the haemoglobin readings meters work in discrete reading of 2 due to difficulty in reading color strips for this test. For the latter example, we updated our rule for haemoglobin to take the discreteness into account (results presented in this paper are with this update) and for the former example, we intend to incorporate some regular field monitoring data in future (if regular field monitoring is economically viable).

For further sanity check, we also analyzed whether there was any effect on the ANMs behavior when being observed versus when she was not. We used the data of ANMs who were observed and who were not observed in 4 weeks from February 2021. Overall, we see that there is not a significant effect from the observers, as shown in the Table~\ref{observer_ef}.

\begin{table}[t]
\centering
\caption{Comparison of the number of ANMs identified by diligent or non-diligent by the AI and the field monitors}
\label{observations}
\begin{tabular}{l|l|c|c|}
\multicolumn{2}{c}{}&\multicolumn{2}{c}{Monitors' observations}\\
\cline{3-4}
\multicolumn{2}{c|}{}& Diligent & Non-diligent \\
\cline{2-4}
\multirow{2}{*}{AI's result}& Diligent & 11 & 6 \\
\cline{2-4}
& Non-diligent & 6 & 13 \\
\cline{2-4}
\end{tabular}
\end{table}
\begin{table}[t]
\centering
\caption{Observer effect analysis}
\label{observer_ef}
\begin{tabular}{p{6cm} p{2cm} p{2cm}  }
\toprule
\bfseries  & \bfseries With observers & \bfseries Without observers \\
\midrule
New score & 1.6797 & 1.7092 \\
Mean of the difference between new score and average past score & -0.1590 & -0.1051 \\
Standard deviation of the difference between new score and average past score & 0.2558 & 0.2150 \\
\bottomrule
\end{tabular}
\end{table}

\section{Discussion and Lessons Learned}
\label{sec:discuss}

Our final product has been through many iterations and prior failed field trails. Here we distill a list of key lesson learned based on our experience.


\begin{itemize} 
\item \textbf{Lesson 1: get continuous feedback from the ground}: Our initial product was not considered useful by the field team. The reasons for these are 
\begin{itemize}
\item In our first iteration we had used six addtional rules provided by the NGO, which provided two clusters with very clear separation provided by three of these six rules. However, the field team rejected our conclusion and it turned out that these three rules are not indicative of diligence. These three rules actually requires the patients to get certain tests done at health centers and report to ANMs, thus, these rules measured behavior of patients rather than ANMs. These six rules were dropped altogether in the final product.
\item We had used a binary classifier to classify diligent and non-diligent ANMs, which had more than 90\% accuracy. However, the field team required a more fine-grained score in order to do the interventions, which we provide now.
\end{itemize}
\item \textbf{Lesson 2: blackbox explanations are futile}: We use blackbox explanation to refer to explanation that do not make sense to the stakeholders (field team of NGO). Our initial attempt at describing past ANM behavior (in first iteration) was in terms of quantitative importance of rules inferred by assigning importance weights to the rules (a common explainablity approach). These descriptions were useless for the NGO team. In response, we developed easier explanation using interpretation of clusters centers at different levels of descriptiveness for different perceptive skills of the stakeholders.
\item \textbf{Lesson 3: use techniques relevant for the problem and not vice versa}: In an intermediate version of our tool, we used a LSTM network to predict the Fuzzy cmeans cluster scores generated with two clusters. Although it had very high accuracy and R2 scores, the initial testing of this released package proved to be not convincing to the field monitors' observations and the fuzzy cmeans provided very poor explanations. In fact, our linear regressor provides very close performance and our strict clusters much better explanation of ANM's behavior.
\item \textbf{Lesson 4: be ready for production environment to change}: Our experience with dropping rules and the discovery of newer non-diligence based on field monitor observations reinforce the fact that the production environment is ever evolving. For this precise reason, our AI system has the flexibility to add/change/delete rules as well as a number of other configurable parameters such as number of clusters.
\end{itemize}

\section{Summary and Future Work}
Our work is in final stages of deployment and will continuously generate behavior explanations and predict scores every month for the ANMs as well as measure their performance in hindsight. Moreover, the data format for tracking pregnancy health used in Rajasthan is consistent with National Health Mission Guidelines followed across India. Thus, this work has the potential to be broadly applied across a health workforce of 250,000 ANMs who care for 25M pregnant women annually. In fact, our  framework of measuring diligence, predicting the same, and providing explanation of CHW behaviors can be applied in any part of the world where CHWs collect data and data collection diligence is an issue. 

Listing a few known limitations in broadly applying our technique, we identify that the model itself will need to be tuned to new data and new CHWs. New domain specific insights from the field such as deviation in GPS points, match rate of entries with beneficiary responses over integrated voice response system (IVRS), and distribution of time stamps during which the app was used on the camp day may all further improve the clustering and prediction; our approach is flexible to take all these into account but the insights require domain expertise. 

\textbf{Ethical considerations}: A future work topic is design of intelligent interventions that are effective in nudging the ANMs to be more diligent in data collection. The design of interventions is a complex social science issue~\cite{sharma2014factors} with ethical considerations. Our interventions planned are not punitive in nature but will focus on encouragement via phone messaging and active on on one engagement with the ANM. Intervention for CHWs on other aspects have shown tremendous promise in addressing related problems~\cite{wells2011community,1571671}.
In the longer term, this work has implications for using pregnancy health data to predict adverse maternal and child health outcomes. By having a diligence score available, data from community health workers can be quantitatively weighted to ensure fidelity of prediction models, which can benefit millions of pregnant women and newborn children. 


\bibliographystyle{acm}
\bibliography{hdps}

\clearpage

\appendix

\section{Additional Results for Baselines}

\begin{figure}
\centering
\includegraphics[width=0.45\textwidth]{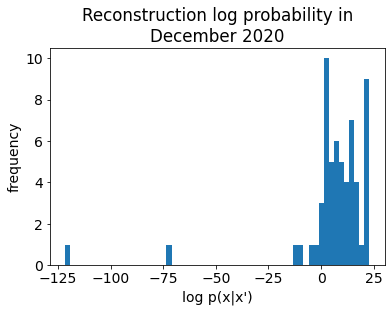}
\includegraphics[width=0.45\textwidth]{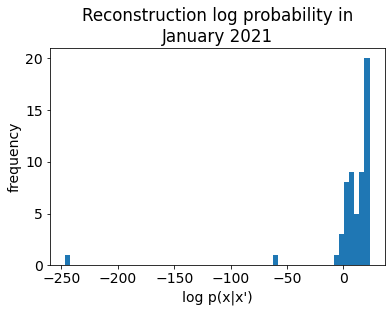}
\caption{Log reconstruction plots using the variational autoencoder}
\label{fig:vae_recons}
\end{figure}

As stated in Section~\ref{sec:results}, we used a variational auto-encoder based anomaly detector as our second baseline to detect non diligent ANMs. Log reconstruction plots for the two months in test data set are available in Figure~\ref{fig:vae_recons}. The ANMs with lower reconstruction log probabilities correspond to anomalies. We chose 0 as the threshold to separate the diligent and non-diligent ANMs. Table~\ref{tab:vae2} shows the percentages of ANMs tagged as non-diligent by the anomaly detector in the test set. These percentages are considered small by our NGO partners. 

\begin{table}
\caption{Percentages of non-diligent ANMs tagged using variational auto encoder based anomaly detector}\smallskip
\label{tab:vae2}
\centering
\begin{tabular}{l c c c }
\toprule
\bfseries  & \bfseries December 2020 & \bfseries January 2020 \\
\midrule
Percentage & 11.67 & 10.53 \\
\bottomrule
\end{tabular}
    
\end{table}

\section{Additional results for the field test} \label{field}

\begin{table}
\centering
\caption{Analysis of the ANMs who are predicted as non-diligent by the ANM, but observed as diligent by the field monitors. (ANM IDs are anonymized)}
\label{tab:anm_analysis}
\begin{tabular}{ l  l p{5cm}  }
\toprule
\bfseries Week & \bfseries ANM ID & \bfseries Non-diligent rule/s\\
\midrule
2 & a & Non-diligent in two short term rules (4th and 5th rules) \\
3 & b & Non-diligent in two short term rules (3rd and 5th rules) \\
3 & c & Non-diligent in three short term rules (3rd, 4th and 5th rules) \\
3 & d & Non-diligent in two short term rules (4th and 5th rules) \\
3 & e & Non-diligent in two short term rules (4th and 5th rules) \\
4 & f & Non-diligent in one short term rule (4th rule) \\
\bottomrule
\end{tabular}
\end{table}

In the field test we see an agreement rate of 67\%. For another 17\% of ANMs, where the AI predicts as non-diligent but the field monitors observe as diligent, we analyze the actual  non-diligence probability vector (corresponding to 11 rules) of each ANM in the observed period. We find that these ANMs were actually non-diligent w.r.t certain rules as shown in the Table~\ref{tab:anm_analysis}, which proves that the AI was correct in predicting these ANMs as non-diligent. Hence, our AI system's accuracy is about 84\%.

\section{Questionnaire} \label{questions}

The field monitors were provided a list of questions relating to the ANM performance (a summary of the questionnaire is given below) and they recorded responses based on their observation.

\begin{itemize}
    \item Whether equipment needed for each test is available.
    \item Whether the ANM conduct each test properly or fills in values without checking or manipulate checked data corresponding to each of the rules. Few example questions related to BP rule recording are as below.
    \begin{itemize}
        \item Does the ANM properly check BP for all women?	
        \item How many women had BP, 120/80?	
        \item Does the ANM enter data without the checking?
        \item Does the ANM enter rounded-off BP values?
        \item Does she record data for few patients and enter "No equipment available" for few other patients?
    \end{itemize}

    \item Whether infrastructure facilities necessary for the health camp are available
\end{itemize}

\end{document}